\documentclass[journal]{IEEEtran}
\usepackage{amsfonts}
\usepackage{amssymb}
\usepackage{amsthm,amsmath}
\usepackage{mathrsfs}
\usepackage{indentfirst}
\usepackage{multirow}
\usepackage{graphicx} %需要使用的宏包
\usepackage{cite}
\ifCLASSINFOpdf
\else
\fi

\begin{document}
\title{TCR-GAN: Predicting tropical cyclone passive microwave rainfall using infrared imagery via generative adversarial networks}

\author{Fan Meng, Tao Song,~\IEEEmembership{Senior Member,~IEEE}, Danya Xu
% <-this % stops a space

\thanks{(Corresponding author: Tao Song)}
% <-this % stops a space

\thanks{F. Meng is with the School of Geosciences, China University of Petroleum, Qingdao 266580, China (e-mail: vanmeng@163.com).}% <-this % stops a space
\thanks{T. Song is with the College of Computer Science and Technology, China University of Petroleum, Qingdao 266580, China (e-mail: tsong@upc.edu.cn).}% <-this % stops a space
\thanks{D. Xu is with Guangdong Laboratory of Marine Science and Engineering(Zhuhai), Zhuhai 519080, China (e-mail:xu-danya@sml-zhuhai.cn)}% <-this % stops a space
\thanks{Manuscript received December 3, 2021; revised xxxx.}}


\markboth{IEEE Geoscience and Remote Sensing Letters}%
{Meng \MakeLowercase{\textit{et al.}}: TCR-GAN: Predicting tropical cyclone passive microwave rainfall using infrared imagery via generative adversarial networks
}



\maketitle

\begin{abstract}
Tropical cyclones (TC) generally carry large amounts of water vapor and can cause large-scale extreme rainfall. Passive microwave rainfall (PMR) estimation of TC with high spatial and temporal resolution is crucial for disaster warning of TC, but remains a challenging problem due to the low temporal resolution of microwave sensors. This study attempts to solve this problem by directly forecasting PMR from satellite infrared (IR) images of TC. We develop a generative adversarial network (GAN) to convert IR images into PMR, and establish the mapping relationship between TC cloud-top bright temperature and PMR, the algorithm is named TCR-GAN. Meanwhile, a new dataset that is available as a benchmark, Dataset of Tropical Cyclone IR-to-Rainfall Prediction (TCIRRP) was established, which is expected to advance the development of artificial intelligence in this direction. Experimental results show that the algorithm can effectively extract key features from IR. The end-to-end deep learning approach shows potential as a technique that can be applied globally and provides a new perspective tropical cyclone precipitation prediction via satellite, which is expected to provide important insights for real-time visualization of TC rainfall globally in operations.

\end{abstract}


% \fi
\IEEEpeerreviewmaketitle

\section{Introduction}
\IEEEPARstart{T}{ropical} cyclones are one of the extreme weather phenomena that cause the heaviest disasters to humans and can cause extreme winds, rainfall, flooding and other disasters. TC originate from the tropical oceans and their fundamental energy comes from the water vapor provided by the ocean\cite{lighthill1994global}, so they carry a lot of water vapor and can bring heavy rainfall, which may lead to extensive flooding, and it is important to estimate the TC rainfall in real time\cite{chen2010overview}.

TC are highly organized weather phenomena where infrared cloud maps are obscured by thick cloud cover and can only measure cloud top features indirectly associated with rainfall\cite{miller2013illuminating}. Passive microwave (PMW) sensors can penetrate the thick cloud cover of TC and detect the thermal radiation of raindrops or the scattering of upwelling radiation caused by precipitation particles\cite{laviola2011183}. Infrared sensors are usually carried by geostationary satellites and thus continuously monitor TC with high temporal-spatial resolution 24 hours a day. Passive microwave sensors are usually carried only by polar-orbiting satellites because of signal loss over long distances, but polar-orbiting satellites can pass the same location only twice a day, so there are severe spatial and temporal constraints through PMW sensors.

There have been related studies to convert IR signals into surface precipitation rates by statistical and machine learning techniques, and the PERSIANN \cite{hsu1997precipitation} algorithm used artificial neural network techniques to establish the relationship between cloud-top bright temperature and surface precipitation rate, and there are numerous subsequent improvements based on this algorithm \cite{sadeghi2019persiann,sadeghi2021persiann}. Deep learning techniques have also proven powerful in infrared precipitation estimation, \cite{tao2018two} developed a stacked noise reduction self-encoder based on a deep neural network oh to give estimates of infrared and water vapor precipitation, and \cite{wang2020infrared} used convolutional neural networks for rainfall estimation.

However, existing studies are directly discussing global precipitation and lack specialized studies on TC, which are highly organized and have significantly different cloud characteristics compared to cloud-free and thin cloud areas\cite{yang2016satellite}. The special characteristics of TC and the catastrophic consequences they can cause make research in this direction of profound importance. GAN trained by a strategy of two networks gaming, have a wide range of applications in the field of computer vision, being applied to video and image generation, where image-to-image translation is a class of visual and graphical problems whose goal is to learn the mapping between input and output images, and has yielded surprising results in directions such as automatic coloring techniques results\cite{isola2017image,zhu2017unpaired}. Considering that the estimated prediction of IR to PMW is essentially an image-to-image translation, GAN is an ideal approach to solve this problem.

In summary, the contributions of this letter can be divided into three items as follows:

(1) We developed a generative adversarial network (TCR-GAN). It attempts to provide direct estimates of PMR from high temporal resolution satellite IR images, providing direct rainfall estimates in the absence of microwave images. To the best of our knowledge, this is the first study dedicated to TC.

(2) We build the benchmark dataset of benchmark (TCIRRP) containing more than 70,000 pairs of paired IR images and passive microwave rainfall to facilitate in-depth studies for the community.

(3) Experimental results show that this end-to-end deep learning approach has a high accuracy and a high potential for real-time precipitation estimation applications during TC.

\begin{figure}[htbp] 
\centering
\includegraphics[width=0.5\textwidth]{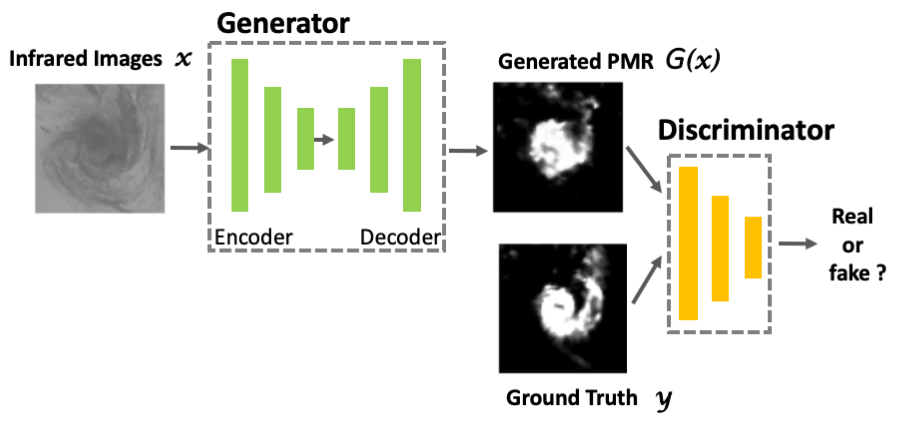} 
\caption{Overview of the TCR-GAN model, consisting mainly of a Unet-like generator and a PatchGAN discriminator.} 
\label{fig_1} 
\end{figure}

\begin{figure*}[htbp]
\centering
\includegraphics[width=6.5in]{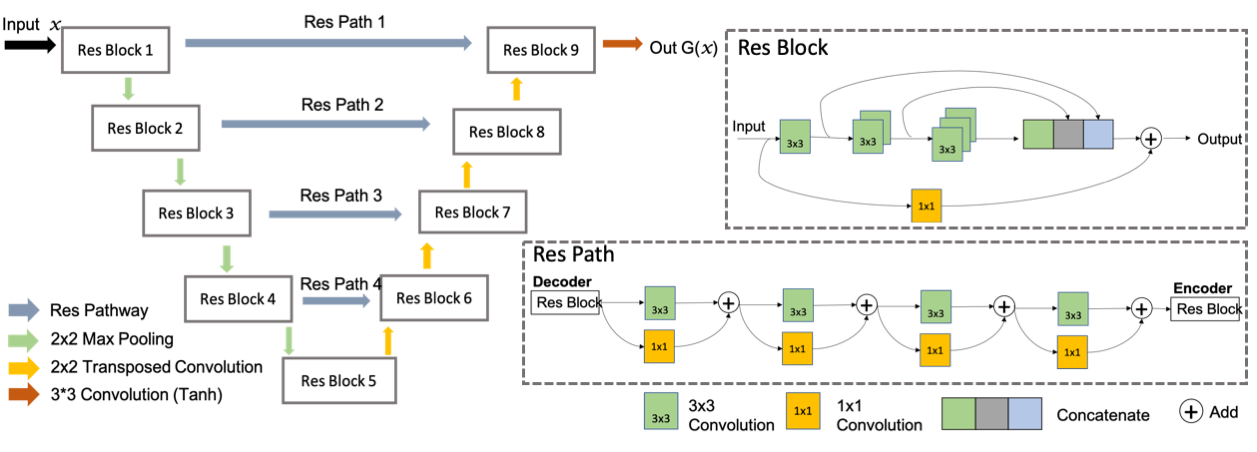}
\caption{Structure details of the Generator, including Res Block and Res Path details.}
\label{fig_sim2}
\end{figure*}

\section{Data and Methods}

%\IEEEpubidadjcol

\subsection{TCIRRP Dataset}

In this study, we constructed the TCIRRP dataset to acquire and process IR channels and PMW channels from the Dataset of Tropical Cyclone for Image-to-intensity Regression (TCIR) dataset, TCIR \cite{chen2018rotation} is designed to help researchers to access satellite images more easily.

The satellite observations involved in TCIRRP data come from two open sources:

(1) GridSat : A long-term dataset of global infrared window brightness temperatures containing the IR channels used in this study using as IR to data from most meteorological geostationary satellites every three hours.

(2) CMORPH \cite{joyce2004cmorph} : CMORPH provides global precipitation analysis at relatively high spatial and temporal resolution, using precipitation estimates derived exclusively from LEO microwave satellite observations, with features transmitted through spatially propagated information obtained exclusively from geostationary satellite IR data.

TCIRRP collects all TC from 2003 to 2017 for which it is possible to obtain a match between the IR channel and the PMW channel, with the center of the TC in the middle.The original resolution of the PMW channel from CMORPH is 1/4 degree lat/lon. In order to unify IR and PMW and to correspond the pixel points one by one, we scale PMW channel about 4 times larger by linear interpolation. The spatial resolution in latitude and longitude is 7°. The size of all images is 201 × 201 points, and the actual distance between each point is about 4 km. There are a few null values in the data, and we interpolate the null values using 0. Based on the statistical values, the IR and PMW channels were normalized separately, and finally more than 70,000 one-to-one pairs of images were obtained. Due to the limitation of computing resources, this study only involved all paired data in 2017, totaling 4549 pairs of images. The specific steps of data processing will be presented in the Experimental section.

\subsection{TCR-GAN}

\subsubsection{Network Architecture}

We propose TCR-GAN to establish the mapping between IR images and PMR. Figure \ref{fig_1} shows the overall architecture of the proposed TCR-GAN, which consists of a generator and a discriminator, trained according to the adversarial strategy. The basic idea is to first generate the predicted PMR image using a generator with a symmetric structure similar to a self-encoder, and then use the discriminator to determine whether the true or false PMR image is the same as the IR and whether it corresponds to each other. The discriminator is then used to determine whether the real or false of PMR images and whether the PMR images correspond to the IR images. Details of the code implementation can be found in the supporting material.

The architecture of the TCR-GAN generator is shown in Figure \ref{fig_sim2}. The construction of the generator is basically the same as that of Pix2Pix\cite{isola2017image}, and we use a modified version of Unet\cite{ronneberger2015u}\cite{ibtehaz2020multiresunet}, introducing Res Block to replace the two convolutional layers of Unet and Res Path to replace the direct jump connection in Unet. 

The results of \cite{szegedy2016rethinking} show that parallelizing 3*3, 5*5, and 7*7 convolution operations can improve the multi-resolution analysis capability of the Unet architecture, and this form can be replaced by a series of smaller 3*3 convolution operations and multiple jump connections, as shown in the upper part of the Res Block image in Figure \ref{fig_sim2}. Instead of having three consecutive convolution blocks using the same number of filters, the Instead of letting the three successive convolution blocks use the same number of filters, the number of filters is gradually increased to cope with the quadratic effect that would exist in successive convolution layers, and this method can greatly reduce the memory requirement. It also introduces a residual connection based on the 1*1 convolution. The 1*1 convolution kernel can increase the nonlinear characteristics while keeping the scale of the feature map constant (i.e., no loss of resolution), which helps construct a deeper network, so that additional spatial information can be effectively added to reduce the problem of gradients disappearing in the Res Block.

Although Unet's shortcut connection between encoding and decoding well preserves the problem of spatial information loss caused by downsampling, it can be noted that the encoder features are lower-level features, while the decoder obtains higher-level features through multi-layer computation. Especially for the first shortcut connection, there is a large semantic gap between these two types of features, so adding some convolutional layers to the shortcut connection can add additional nonlinearities and construct some low-level features, which can compensate the problem of large semantic gap between the encoder and decoder to a certain extent. The constructed shortcut connection res block is shown in Figure \ref{fig_sim2}, which connects the encoder and decoder through several consecutive residual connections composed of 1*1 and 3*3 convolutional kernels, which can maintain more feature information and add additional feature information. All the convolutional layers in the whole generator network except the output layer are activated by the ReLU activation function with batch normalization, and the output layer we are activated by the tanh activation function.

\begin{figure}[!t]
\centering
\includegraphics[width=3.5in]{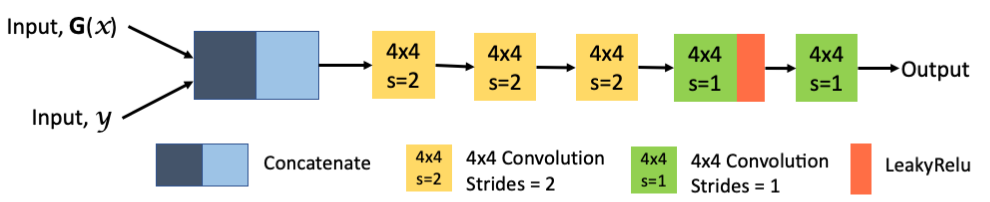}
\caption{Construction details of the Discriminator.}
\label{fig_sim3}
\end{figure}

In order to make better judgments on the localization of the image, the discriminator of TCR-GAN uses the discriminator construction of PatchGAN, and the construction of the discriminator in this study is shown in Figure \ref{fig_sim3}. While the general GAN discriminator simply outputs a ture or fasle vector to represent the evaluation of the whole image, PatchGAN finally outputs an $N*N$ matrix through multiple convolutional layers, trying to classify each element of the $N*N$ matrix as True or False. each element actually represents a larger perceptual field in the original image, i.e., it restricts the attention in the patch corresponding to different scales of localization in the original graph. Such a discriminator effectively models the image as a Markov random field, assuming independence between pixels separated by more than a patch diameter.

\subsubsection{Loss function}

The optimization objective of TCR-GAN contains 2 parts, one is the optimization objective of cGAN and the other is the L1 distance, which is used to constrain the difference between the generated PMR image and the real image, with the aim of reducing the blurring of the generated image. L1 loss can recover the low frequency part of the image, while GAN loss can recover the high frequency part of the image, and the formula is shown as follows:

$$
G^{*}=\arg \min _{G} \max _{D} \mathcal{L}_{c G A N}(G, D)+\lambda \mathcal{L}_{L 1}(G)
$$

where G tries to minimize the objective while D tries to maximize it, and lanmda is the empirical value, which we set to 100 by default in this study. Specifically, the optimization objective of cGAN is given by the following equation, with z denoting random noise:

$$
\mathcal{L}_{c G A N}(G, D)= \mathbb{E}_{x, y}[\log D(x, y)]+\\
 \mathbb{E}_{x, z}[\log (1-D(x, G(x, z))]
$$

For the image translation task, a lot of information is actually shared between the input and output of G. In the task of this study, information such as TC location and size is shared between the input and output. Thus, to ensure the similarity between the input and output images and to constrain the difference between the generated PMR image $G(x, z)$ and the real PMR image y, L1 loss is also incorporated, which is shown in the following equation:

$$
\mathcal{L}_{L 1}(G)=\mathbb{E}_{x, y, z}\left[\|y-G(x, z)\|_{1}\right].
$$

\section{Experiments and Results}

\subsection{Experiments Setups and Evaluation Metrics}

This study only involves all the 2017 data in TCIRRP we constructed, totaling 4,579 pairs of images, and the data are arranged in chronological order, due to the possible large similarity of adjacent TC images, and in order to keep the training and test sets relatively independent, we take the first 4,000 pairs of images as the training images and all the remaining images as the test set. To prevent overfitting, we adjust the images to a larger height and width, and perform data enhancement methods such as random cropping and horizontal mirroring and normalization, the specific operations and data can be found in our open source code.

\begin{figure*}[htbp]
\centering
\includegraphics[width=5in]{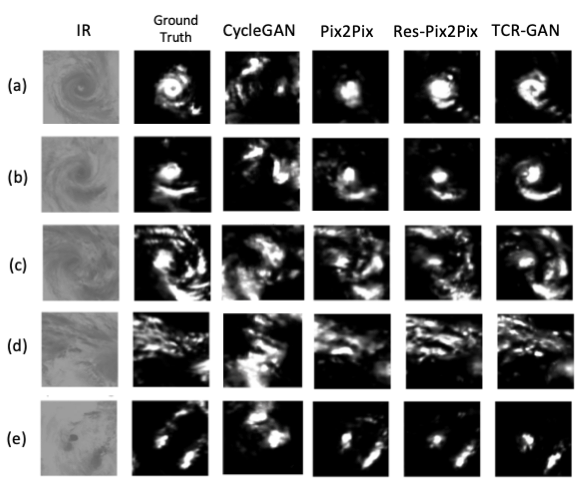}
\caption{The prediction results of 5 typical TC with different characteristics cover the various cycles and development states of TC development.}
\label{fig_results}
\end{figure*}

To evaluate the performance of the proposed prediction model and to compare it with other models, four commonly used statistical metrics were used. (1) Peak Signal-to-Noise Ratio (PSNR), which is the ratio of the energy of the peak signal to the average energy of the noise. (2) The Root Mean Square Error (RMSE), which can assess the difference between the forecast and actual values. (3) Pearson correlation coefficient (CC) that can describe the level of linear correlation between the PMR estimate and the actual value. (4) Structural SIMilarity (SSIM) of luminance, contrast, and structure can be evaluated to quantitatively describe the performance of the proposed model. Among them, the PSNR and SSIM are calculated as follows:
$$
P S N R=10 \times \lg \left(\frac{255^{2}{N}}{{\sum_{n=1}^{N}\left(x^{n}-y^{n}\right)}}\right),
$$

$$
\operatorname{SSIM}(x, y)=\frac{\left(2 \mu_{x} \mu_{y}+c_{1}\right)\left(2 \sigma_{x y}+c_{2}\right)}{\left(\mu_{x}^{2}+\mu_{y}^{2}+c_{1}\right)\left(\sigma_{x}^{2}+\sigma_{y}^{2}+c_{2}\right)},
$$

where $N$ is the size of image; $x^{n}$ and $x^{n}$ are the $nth$ pixels of original image $x$ and processed image $y$; $\mu_{x}$ and $\mu_{y}$ are the averages of $x$ and $y$; $\sigma_{x}^{2}$ and $\sigma_{y}^{2}$ are the variance of $x$ and $y$; and $\sigma_{x y}$ is the covariance of x and y.

Since the task of this study is essentially an image-to-image translation task, our model is compared to the advanced model CycleGAN \cite{zhu2017unpaired}. Another reason for using CycleGAN is that the method uses non-paired data to learn the mapping of two types of images, and we try to verify whether the task can be done with non-paired data. Since TCR-GAN is essentially a modified version of Pix2Pix, Pix2Pix \cite{isola2017image}was also applied to the comparison. Further, we modified the Unet architecture in the generator by replacing the two-layer convolution with a res block, which we named Res-Pix2Pix, an experimental setup that verifies the role of the Res Block with the native Pix2Pix and also serves as an ablation experiment to verify the role of the Res Path.

All models are trained with 100 epochs using the Adam optimizer, with the Adam optimizer momentum parameter $\beta1 = 0.5$, the learning rate of both the generator and the discriminator set to 0.0002, and the batch size set to 1. All models in this study are implemented using tensorflow. Our experimental and validation environments are as follows: Intel Core i9-9900K CPU with Geforce RTX 3090 GPU, 128G RAM, Ubuntu 18.04, and our model training time is close to 2.5 hours in the dataset involved in this study.
\subsection{Results}

From Table \ref{tab:my-table}, which shows the quantitative evaluation results of TCR-GAN and the comparison models on the test set, in general, the results of TCR-GAN outperform the other models in all performance metrics, Cyclegan performance is much weaker than the other three models. Comparing the results of the other three models, it can be noted that the performance is improving with the gradual improvement of Pix2Pix, it should also be noted that The improvement of the four indicators is not very large, and we believe that the important reason is due to the fact that the value of no rainfall, that is, the 0 value, accounts for the vast majority of the image. Through our qualitative comparison of the test set we found that although the difference in the indicators is small, in practice there is still a large gap.

Figure \ref{fig_results} shows five typical examples of predictions in the test set. All predictions for the test set can be found in the supporting material. On the whole, TCR-GAN has more similarity with ground truth than the other four models and is able to capture the key information better, but it is worth noting that although there is a great similarity in structure and some deviation in image brightness in some regions, i.e., PMR regions can be accurately predicted, in terms of intensity prediction, the performance in details is not very accurate. Interestingly, CycleGAN learns features that cleverly avoid the essential features and learn complementary information because they are not paired for training. Specifically, Figure \ref{fig_results}(a) shows an example of shaped with eyes, where TCR-GAN captures the typhoon eyes and spiral rain bands. Further, Figure \ref{fig_results}(b) shows a TC containing significant spiral rain bands, which are captured relatively accurately by TCR-GAN. Figure \ref{fig_results}(c) and Figure \ref{fig_results}(d) show TC with complex situations and TC showing a high degree of dispersion, respectively. While other models lose much critical features, our proposed model has a better capture of critical information such as spiral rainbands. Figure \ref{fig_results}(e) shows the model's prediction of a dissipating TC, the critical features is well captured by each model, but the weak information in the upper right corner of the figure is selectively ignored by the model as redundant information.

Through the above qualitative and quantitative evaluation, it can be concluded that TCR-GAN learns the mapping features of IR and PMW images well by training, has robustness and generalization, and our model can predict key regions such as spiral rain bands more finely. However, the performance in details is not very accurate, we assume that it is due to the limitation of computation and only 1 year of data has been used for the study, if it can use the whole TCIRRP data, we think there will be a great improvement in the detailed and intensity forecasts.

\begin{table}[]
\centering
\normalsize
\caption{Comparison of evaluation metrics of TCR-GAN and comparison models.The best results are marked in bold.}
\label{tab:my-table}
\begin{tabular}{lcccc}

\hline
Models      & PSNR            & RMSE           & CC              & SSIM           \\ \hline
CycleGAN    & 9.781           & 7.433          & 0.096          & 0.397          \\
Pix2Pix     & 14.080          & 6.861          & 0.596          & 0.530          \\
Res-Pix2Pix & 14.376          & 6.848          & 0.623          & 0.542          \\
TCR-GAN     & \textbf{14.480} & \textbf{6.705} & \textbf{0.637} & \textbf{0.550} \\ \hline
\end{tabular}
\end{table}

\section{Conclusion}

In this letter, we develop a GAN-based algorithm TCR-GAN to directly predict microwave rainfall images using IR images for the specificity of TC remote sensing images.We provide a benchmark dataset TCIRRP. To verify the effectiveness of TCR-GAN, the model is trained on 4000 pairs of matched remote sensing images, and nearly 600 images were tested, and the results showed that TCR-GAN can accurately and effectively extract the key features of microwave rainfall from IR images. During the testing period, TCR-GAN obtained good statistical performance performance based on RMSE,CC,PSNR,SSIM.

Overall, TCR-GAN offers new possibilities for PMR real-time estimation for tropical cyclones and provides an effective method for directly predicting rainfall intensity and area during tropical cyclones using infrared images, which is expected to be an effective auxiliary forecasting tool for operational forecast centers. However, there are limitations in this study; the limits of the accuracy of the algorithm depend on the accuracy of the PMR products used, and the algorithm uses image-to-image conversion for forecasting and does not provide accurate quantitative precipitation, so future work is to train IR to match precipitation values. Since the inherent limitations of IR instruments also lead to inaccurate estimates, future work could use multi-channel data for forecasting, such as the water vapor channel and the visible channel.

% or


% \appendices

% % you can choose not to have a title for an appendix
% % if you want by leaving the argument blank
% \section{}

\section*{Acknowledgment}

The authors would like to thank Boyo Chen of National Taiwan University for providing the open source TCIR dataset, which is used as the source data for our construction of TCIRRP.

\ifCLASSOPTIONcaptionsoff
  \newpage
\fi



%\bibliographystyle{IEEEtran}
\bibliographystyle{IEEEtran}
\bibliography{IEEEexample}



% \end{thebibliography}


% \end{IEEEbiography}


% % insert where needed to balance the two columns on the last page with
% %\newpage



%\vfill


\end{document}